# Recursive Filter for Space-Variant Variance Reduction

Alexander Zamyatin[1]

*Abstract–* The proposed method reduces non-uniform sample variance to the pre-determined target level. The space-variant filter can equalize variance of the non-stationary signal, or vary filtering strength based on image features, such as edges, etc., as shown by applications in this work. The proposed approach computes the variance reduction ratio at each point of the image, based on the given target variance. Then, a space-variant filter with matching variance reduction power is applied. A mathematical framework of atomic kernels is developed to facilitate stable and fast precomputing of the filter bank and kernel indexing. Recursive formulation allows using small kernel size, which makes the space-variant filter more suitable for fast parallel implementation. Despite the small kernel size, the recursive filter possesses strong variance reduction power. Filter accuracy is measured by the variance reduction against the target variance; testing demonstrated high accuracy of variance reduction of the recursive filter compared to the fixed-size filter. The proposed filter was applied to adaptive filtering in image reconstruction and edge-preserving denoising.

## 1 Introduction

We propose a new method for space-variant image filtering, where the filter kernel varies with pixel position in the image. Applications of space-variant filters include adaptive filtering, edge-preserving denoising, texture mapping [1], and foveated coding [2]. Due to variable kernel size, Fast Fourier Transform (FFT) convolution is not applicable for space-variant filtering, and the filter has to be implemented in spatial domain. Therefore, computation time is the main concern for space-variant filters, especially for applications requiring larger kernel sizes, such as adaptive filtering or foveated coding. Note that the direct convolution method becomes prohibitively slow when the kernel is computed for every image pixel.

Previously published approaches for fast implementation of spatial domain convolution can be classified in the following groups: *Integral methods* [3] [4] [5] [6], *Cosine integral* methods [7], [8], and *Recursive methods* [9], [10], [11]. Crow's summed-area tables [3] used for fast calculation of local sums in box filtering [12], were later generalized by Heckbert [4] for *repeated integration*, a method that computes piece-wise polynomial kernels of degree $n$; the drawback of this approach is that the filter strength cannot be chosen arbitrarily, since $n$ is an integer. Cosine integral methods approximate the Gaussian kernel as a weighted sum of truncated Cosine functions. Recursive methods approximate a Gaussian filter as a combination of causal and anti-causal filters and require both forward and backward recursions. Even though they compute Gaussian kernels with high accuracy, both *cosine integral* and *recursive* methods do not lend themselves easily for space-variant filtering.

The main idea to accelerate space-variant filtering is to precompute a filter bank that can be used to generate kernels at each pixel. Popkin at al [13] presented a computationally efficient algorithm for smoothly space-variant Gaussian image blurring, that uses a specialized filter bank with optimal filters computed through principal component analysis. This filter bank approximates space-variant Gaussian blurring to arbitrary degree of accuracy. The price to pay for high-accuracy Gaussian approximation is a large kernel size of 81x81 pixels and algorithmic complexity.

In this work we select a recursive approach to implement the space-variant filter with small-footprint. Even though the method is developed for arbitrary kernel size, practical applications use 3x3 recursive filters for variance reduction up to a factor of 100. In special cases, when higher variance reduction is needed, larger size filters can be employed. Similar to [13], we precompute a filter bank, so that the convolution kernels do not have to be computed at each image point; they are selected from the filter bank and kernels can be re-used many times.

We introduce the notion of *target variance* and apply the recursive filter to achieve the target variance at each point. Target variance is given by the application in hand. Thus, to evaluate the accuracy of the proposed method, the measured variance in the filtered image is compared to the target variance. This accuracy metric based on target variance is different from the prior art, where accuracy is typically measured against an *exact* Gaussian filter. While our filter design is based on Gaussian kernels, it is not our goal to approximate Gaussian kernel as accurately as possible.

This work is organized as follows. In Section 2 we present the mathematical framework and introduce a concept of *atomic kernels* to simplify computation of the recursive filter. Section 3 describes implementation of the proposed approach and includes a baseline fixed-size algorithm and the proposed recursive algorithm. A special subsection is dedicated to closed form equations for 3x3 practical recursive filters. Section 4 demonstrates some applications of the proposed filter, Adaptive filter and edge-preserving image filter.

## 2 PROPOSED METHOD

### 2.1 Notations

Consider an image $f: \mathbb{Z}^2 \to \mathbb{R}$ defined on a rectangular domain $I = [0, W] \times [0, H]$. Image component is denoted by $f_i$, with 2D index $i \in I$. Consider a kernel $h: \mathbb{Z}^2 \to \mathbb{R}$, defined in the square neighborhood $\Omega = [-L, L] \times [-L, L]$. The coefficients of the kernel are denoted $c_j$, with 2D index $j \in \Omega$. The filter coefficients must satisfy the following conditions:

$$\sum_{j \in \Omega} c_j = 1, \quad c_j \geq 0 \text{ for all } j \text{ in } \Omega. \quad (1)$$

The above conditions force $c_j \leq 1$ for all $j$ in $\Omega$. With this notation, the convolved image, denoted $\hat{f}$, is given by:

$$\hat{f}_i = [f * g]_i = \sum_{j \in \Omega} c_j f_{i-j}, \quad (2)$$

where "*" symbol represents the convolution operator.

Denote the variance of $f$ at each point in $I$ by $v_i$, and the target variance by $v_T$. The scope of this work is limited to the variance reduction applications, that is, $v_T \leq v_i$ for all $i$. With this notation, we introduce the *Variance Reduction Ratio* (VRR) as:

$$q_i = \frac{v_i}{v_T}. \quad (3)$$

Next, let $\xi_i$ be independent and identically distributed (iid) random variables. We define the *Variance Reduction Power* (VRP) of the filter $h$ as:

$$p_h = \frac{V[\xi]}{V[\xi * h]}, \quad (4)$$

where $V[\cdot]$ is the variance operator. We show that VRP can be computed from the kernel coefficients, according to the following lemma.

Let $\xi_i$ be iid random variables defined on a rectangular domain $I$ in $\mathbb{Z}^2$. Let $h = \{c_j\}$, $j \in \Omega$, be a 2D low-pass filter satisfying conditions (1). The variance reduction power of $h$ is given by:

$$p_h = \left( \sum_{j \in \Omega} c_j^2 \right)^{-1} \quad (5)$$

Proof:

$$V[\xi * h]_i = V\left[ \sum_{j \in \Omega} c_j \xi_{i-j} \right] = \sum_{j \in \Omega} V[c_j \xi_{i-j}]$$
$$= \sum_{j \in \Omega} c_j^2 V[\xi_{i-j}] = V[\xi] \sum_{j \in \Omega} c_j^2.$$

Here we used variance properties $V[A + B] = V[A] + V[B]$, and $V[\alpha X] = \alpha^2 V[X]$. Also, since $\xi_i$ is iid, $V[\xi_i] = V[\xi]$ for all $i$. It can shown that for square symmtric kernels computing the VRP (5) can be simplified:

$$p_h = \frac{1}{(\text{trace}(h))^2}, \quad (6)$$

where the kernel trace is defined as the sum of the elements on its main diagonal.

Now that we introduced the key concepts of VRR and VRP, we can summarize our approach:

1. Given the input and target variances of the imaging task, compute the desired VRR at each sample.
2. Given the VRR at a given sample, construct a filter with VRP = VRR.
3. Apply this filter in the neighborhood of the sample to achieve the desired variance reduction.

Next, we will discuss how to construct a filter with a desired VRP. For an efficient implementation, we pre-compute a filter bank $H$, that spans all possible values of VRP, from least to largest. Filter bank eliminates the need to calculate kernels on the fly and allows reusing kernels at different points. For fast addressing kernels in the filter bank, the kernel index shall be directly proportional to the VRP. Also, we need to determine the smallest and largest VRP values of kernels in the filter bank, $p_{\min}$ and $p_{\max}$.

The value $p_{\min}$ is set to 1, and it corresponds to the case when no variance reduction is applied, that is, a kernel $h_0$ having the property that $f * h_0 = f$ for any image $f$. Clearly, $h_0$ is the *delta* kernel:

$$h_0: c_{j_1 j_2} = \begin{cases} 1, & j_1 = j_2 = 0 \\ 0, & \text{otherwise} \end{cases} \quad (j_1, j_2) \in \Omega \quad (7)$$

The value $p_{\max}$ depends on the kernel size ($K = 2L + 1$), since larger kernels have more power for variance reduction. For a given $K \times K$ kernel size, the largest variance reduction is provided by the *box kernel*:

$$h_{N-1}: c_j = \frac{1}{K^2}, j \in \Omega \quad (8)$$

It follows from (5) that $p_{\max} = K^2$ (for a rectangular $K_1 \times K_2$ kernel, we have $p_{\max} = K_1 \times K_2$).

The filter bank kernels continuously span all VRP values from $p_{min}$ to $p_{max}$. Next, we discuss how to compute a kernel with arbitrary VRP, $p_{min} < p < p_{max}$. The most natural choice is to find a Gaussian kernel with given VRP. Recall the equation of the Gaussian kernel $g_\sigma = \{g_j\}$ centered at (0, 0) and defined on $\Omega$:

$$g_j = \frac{\exp\left(-\frac{\|j\|_2^2}{2\sigma^2}\right)}{\sum_{i\in\Omega}\exp\left(-\frac{\|i\|_2^2}{2\sigma^2}\right)}, \quad j \in \Omega \quad (9)$$

where the $\ell_2$ norm is defined by $\|j\|_2^2 = (j_1^2 + j_2^2)$. The width of the Gaussian kernel is determined by the parameter $\sigma$. As $\sigma$ increases and the kernel becomes wider, its VRP increases as well, that is, VRP is a non-linear, monotonically increasing function of $\sigma$ (see Figure 1). In general, however, there is no closed form equation between $\sigma$ and the kernel VRP. Another difficulty is to find values of $\sigma$ corresponding to $p_{min}$ and $p_{max}$. While we can select very small $\sigma$ (e.g., $\sigma_{min} = 0.2$) that produces the delta kernel, there is no value of $\sigma_{max}$ that can produce the box kernel (box kernel is the limiting case of the Gaussian kernel (9) for $\sigma \to \infty$). With sufficiently large $\sigma_{max}$ we can make a kernel with VRP close to $p_{max}$; however, the value of $\sigma_{max}$ depends on the kernel size, as shown in Figure 1.

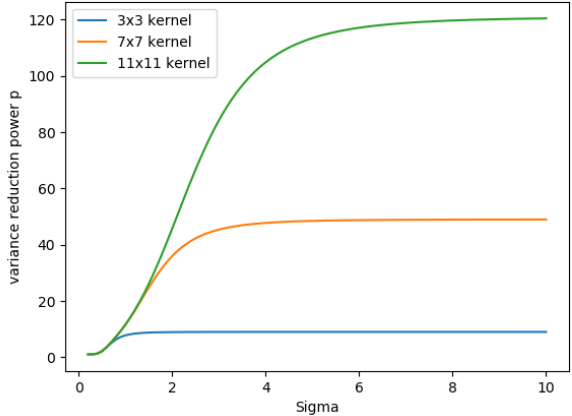

**Figure 1**. Variance reduction power of the Gaussian kernel as a function of $\sigma$ and fixed kernel size. As $\sigma$ increases, VRP approaches its limiting value depending on the kernel size, $p_{max} = K^2$.

To overcome these issues of standard formulation of Gaussian kernels in terms of $\sigma$ (9), we propose the concept of *atomic kernels*.

## 2.2 Atomic kernels

Atomic kernel $A_L$ of size $K \times K$ ($K = 2L + 1$), with parameter $a$, $0 \le a \le 1$, is defined by the self outer product of its generating kernel $U_L$ given by:

$$A_L = U_L \otimes U_L, \quad U_L = \{a^{l^2}\}, \quad l \in [-L, L] \quad (10)$$

For example, $U_1 = (a, 1, a)$, $U_2 = (a^4, a, 1, a, a^4)$, $U_3(a) = (a^9, a^4, a, 1, a, a^4, a^9)$, etc. The size parameter $L$ is a meta-parameter that depends on the imaging task. Examples of 3x3 and 5x5 kernels are shown below:

$$A_1 = U_1 \otimes U_1 = \begin{bmatrix} a^2 & a & a^2 \\ a & 1 & a \\ a^2 & a & a^2 \end{bmatrix},$$

$$A_2 = U_2 \otimes U_2 = \begin{bmatrix} a^8 & a^6 & a^4 & a^6 & a^8 \\ a^6 & a^2 & a & a^2 & a^6 \\ a^4 & a & 1 & a & a^4 \\ a^6 & a^2 & a & a^2 & a^6 \\ a^8 & a^6 & a^4 & a^6 & a^8 \end{bmatrix} \quad (11)$$

Note that the atomic kernel with parameter $a$ corresponds to the truncated Gaussian kernel with parameter $\sigma_a$ given by:

$$\sigma_a = \frac{1}{\sqrt{-2\ln a}}. \quad (12)$$

To satisfy the normalization condition (1), $A_L$ needs to be divided by its sum, $\|A_L\|_1$, where the $\ell_1$ kernel norm $\|\cdot\|_1$ is defined as the sum of its elements (note that all kernel elements are non-negative by definition). Atomic kernels have the property that their sum can be computed fast as the square of the sum of elements along the middle row (or its generating kernel):

$$\|A_L\|_1 \equiv \sum_{j\in\Omega} A_j = \left(\sum_{l\in[-L,L]} a^{l^2}\right)^2 = \|U_L\|_1^2, \quad (13)$$

From (6) VRP of a normalized atomic kernel can be computed by adding elements along the main diagonal:

$$p_L(a) \equiv p[A_L] = \left(\frac{1}{\|A_L\|_1}\sum_{l\in[-L,L]} a^{2l^2}\right)^{-2}$$

$$= \frac{\left(\sum_{l\in[-L,L]} a^{l^2}\right)^4}{\left(\sum_{l\in[-L,L]} a^{2l^2}\right)^2} = \frac{\|U_L\|_1^4}{\|U_L\|_2^2} \quad (14)$$

Advantage of the atomic kernels for constructing a filter bank is that they use a parameter with fixed bounds, i.e., $0 \le a \le 1$. Note that the atomic filter bank contains the box kernel by design.

Next, we need to compute a normalized atomic kernel (i.e., find $a$), given the desired VRP $p \in [p_{min}, p_{max}]$. This requires solving equation (14) for $a$, which involves a high-order polynomial, and analytic solution is not feasible in general (for $K > 3$), and therefore we employ a numerical method to find the kernel parameter $a$. Analytic solution can be found for the simplest $K = 3$ case, where VRP simplifies to:

$$p_1(a) = \frac{(1 + 2a)^4}{(1 + 2a^2)^2}, \quad (15)$$

and the 4th order polynomial equation can be reduced to quadratic equation and solved analytically:

$$a(p) = \begin{cases} 0.25, & p = 4 \\ \dfrac{-2 + \sqrt{2t(3-t)}}{2(2-t)}, t = \sqrt{p}, & \text{otherwise} \end{cases}, (16)$$

for $1 \leq p \leq p_{max}$. The details of numerical solution for general cases $K > 3$ are given in the next section.

## 3 NUMERICAL IMPLEMENTATION

### 3.1 Fixed size algorithm

For numerical solution of (14), we pre-compute a lookup table $P(a)$ for an oversampled set of $a$ values (e.g., 1000 samples), $0 \leq a \leq 1$. Then, for a filter bank index $p$, we find the corresponding value of $a_p$ from $P(a)$ table by inverse lookup.

---
**Algorithm 1A**: Filter bank for fixed-size filter

1. Set $p_{min} = 1$, $p_{max} = K^2$.
2. Pre-compute a lookup table $P(a)$ for $0 \leq a \leq 1$ per (14).
3. For discrete values of $p \in [p_{min}, p_{max}]$ :
   3.1. Find $a_p = argmin(|P(a) - p|)$ (inverse table lookup).
   3.2. Compute the filter bank kernel with VRP of $p$: $H[p] = A_L(a_p)$ (equation (10)).
4. Output $H$
---

Armed with the pre-computed filter bank, we introduce the fixed-size space-variant variance reduction algorithm:

---
**Algorithm 1B**: Fixed-size variant filter

1. Read input arrays $f$ and $q$.
2. Loop over pixels $i \in I$:
   2.1. Given $q_i$, select $p = q_i$.
   2.2. Check if $p > p_{max}$, restrict $p$ to $p_{max}$. This is an inherent limitation of this algorithm.
   2.3. Given $p$ select the kernel $h_p = H[p]$ from the filter bank.
   2.4. Compute convolved value $\hat{f}_i$ by equation (2).
3. Output $\hat{f}$.
---

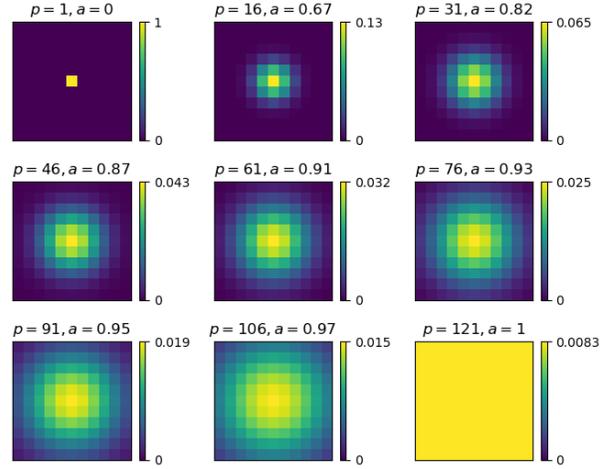

**Figure 2**. Example of the filter bank with 11x11 kernels, indexed by VRP. A representative selection of kernels (9 out of 121) is shown.

An example of a 11x11 filter bank is shown in Figure 2. In the next section, we test this algorithm with computer generated data.

#### 3.1.1 Test 1: Gaussian noise samples

In this test we use a Gaussian random number generator to make $N_s = 200$ noise samples. Each sample has 128x128 pixels, and we use the inner 100x100 block to compute variance. Each 2D sample is generated with zero mean and variance set to $n$, where $n \in [1, N_s]$ is the sample index. That is, expected sample variance is $E[v_n] = n$, but the measured noise variance $v_n$ varies due to the randomness of noise generation. With the input variance $v_n$ varying roughly from 1 to 200, we set the target variance to a fixed value $v_T = 1$. We compare two filter banks of sizes 7x7 and 11x11. Figure 3 plots the measured sample variance $v_n$ and the filtered sample variance $\hat{v}_n$. To improve accuracy of the variance measurements, we repeat this test 100 times, and average the results. These plots are shown in Figure 4.

Figures 3-4 demonstrate the accuracy of the proposed approach. For both cases, 7x7 and 11x11, filtered sample variance is very close to the desired target variance, up to the limit of maximum VRP. The main issue of the fixed-size approach is that the kernel VRP is limited by the kernel size, as shown in Figure 4. This prompts selection of larger size kernels, which greatly increases computational burden, especially if majority of work is done by smaller-size kernels. In the next section, we discuss a recursive approach that overcomes issues stated above.

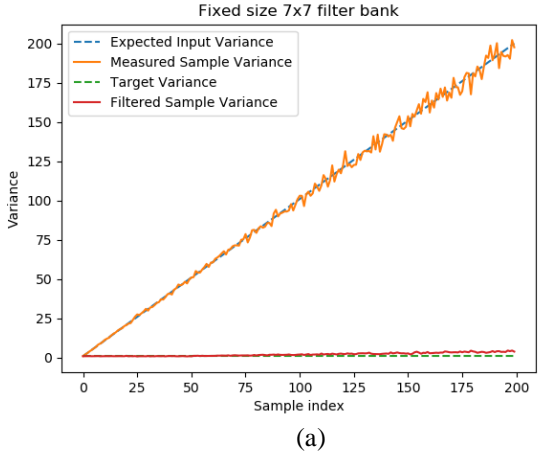

(a)

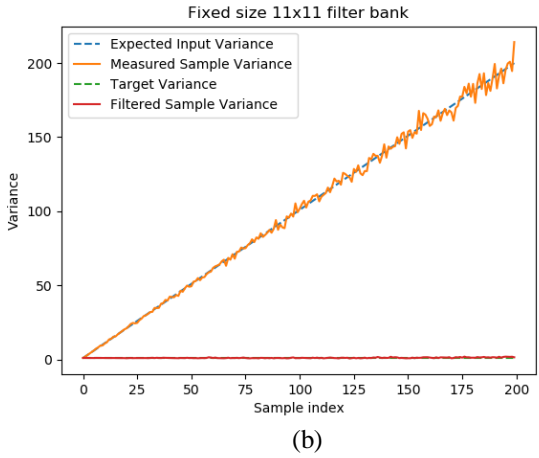

(b)

**Figure 3**. Test 1: Variance reduction of Gaussian noise samples with fixed size kernels. (a) 7x7 filter bank; (b) 11x11 filter bank.

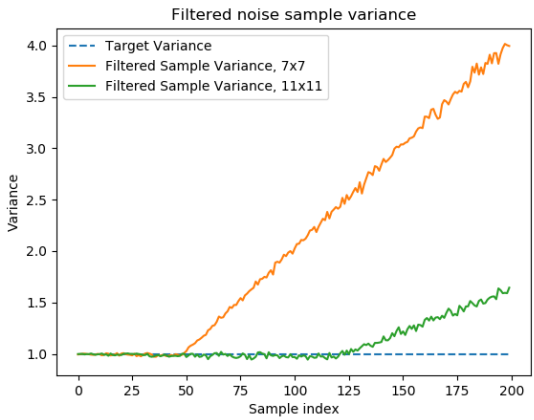

**Figure 4**. Filtered sample variance for two selected cases, 7x7 and 11x11 filter banks, compared to the target variance baseline.

## 3.2 Recursive Implementation

As we discussed above, of all kernels in the filtering chain, the box kernel provides the maximum VRP. Therefore, the maximum VRP at iteration $n$ is given repeatedly applying the box kernel $n$ times. As discussed in the Introduction, the repeated box filter was proposed by Heckbert [6], who also proposed a term *selective image filter*. However, the problem of this approach is that VRP cannot be chosen arbitrary. In the proposed recursive implementation, we improve the repeated box filter algorithm to achieve the desired variance reduction. Denote by $b$ the box kernel defined on $\Omega$ (8), and by $b_n$ the repeated box kernel :

$$b_n = \underbrace{b * \ldots * b}_{n \text{ times}}$$

For completeness, the set of $b_n$ can be augmented with the delta kernel, i.e., $b_0 = h_0$ (7).

The main difficulty of recursive implementation is that successive application of a convolution kernel does not bring the same variance reduction power as the first iteration, that is $p(h_1 * h_2) \neq p(h_1)p(h_2)$. Table 1 lists maximum variance reduction with number of iterations ($p_{\max}^{(n)}$) for 3x3, 5x5, and 7x7 box kernels.

**TABLE 1**. MAXIMUM VARIANCE REDUCTION DEPENDING ON THE NUMBER OF ITERATIONS.

| Iterations | 3x3 filter | 5x5 filter | 7x7 filter |
|---|---|---|---|
| 1 | 9.00 | 25.00 | 49.00 |
| 2 | 18.17 | 54.07 | 108.03 |
| 3 | 26.73 | 79.63 | 158.97 |
| 4 | 35.13 | 104.76 | 209.20 |
| 5 | 43.50 | 129.87 | 259.41 |
| 6 | 51.87 | 154.98 | 309.64 |
| 7 | 60.24 | 180.10 | 359.88 |
| 8 | 68.62 | 205.22 | 410.12 |

From Table 1, the total variance reduction at $n$ iterations can be approximated by a simple model:

$$q = n\, p_{\max} + p_{\text{last}}, \qquad (17)$$

from which the number of iterations can be easily estimated, $n = \left\lfloor \frac{q}{p_{\max}} \right\rfloor$. Here $\lfloor \cdot \rfloor$ notation represents the integer floor function. Here $p_{\text{last}}$ is the VRP at the last iteration, and it can be approximated by $p_{\text{last}} = q - np_{\max} = q \bmod p_{\max}$. In this work, however, we describe an exact approach to perform recursive filter to achieve the target variance reduction. In terms of the VRP definition in Section 2.1, the VRP of the filter $h$ at iteration $n$ is defined by:

$$p_h^{(n)} = \frac{V[\xi]}{V[\xi * h * b_n]}. \qquad (18)$$

This is consistent with non-iterative definition (4) with $n = 0$. Note that

$$p_{\max}^{(n)} = \frac{V[\xi]}{V[\xi * b_{n+1}]}. \quad (19)$$

We also define the *incremental Variance Reduction Power* (iVRP) of the filter $h$ at iteration $n$ as:

$$r_h^{(n)} = \frac{V[\xi * b_n]}{V[\xi * h * b_n]}. \quad (20)$$

Note that given iVRP $r_h^{(n)}$ we can easily compute the full VRP at iteration $n$:

$$p_h^{(n)} = p_{\max}^{(n-1)} r_h^{(n)}, \quad (21)$$

Table 2 shows the maximum incremental variance reduction at each iteration, $r_{\max}^{(0)} = p_{\max}^{(0)}$, $r_{\max}^{(n)} = \frac{p_{\max}^{(n)}}{p_{\min}^{(n)}}$, where $p_{\min}^{(n)} = p_{\max}^{(n-1)}$. Note that each successive iteration has smaller iVRP than the previous iteration.

Atomic kernel at iteration $n$ is defined as $A_L^{(n)} = U_L^{(n)} \otimes U_L^{(n)}$, with the generating 1D kernel $U_L^{(n)}(a)$ given by: $U_L^{(n)}(a) = b_n^1 * U_L(a)$, where $b_n^1$ is the 1D box kernel repeated $n$ times. Similar to (14), we can compute VRP of the atomic kernels at iteration $n$:

$$p_L^{(n)}(a) = \frac{\left\|U_L^{(n)}(a)\right\|_1^4}{\left\|U_L^{(n)}(a)\right\|_2^2} \quad (22)$$

**TABLE 2**. MAXIMUM INCREMENTAL VARIANCE REDUCTION AT EACH ITERATION.

| Iterations | 3x3 filter | 5x5 filter | 7x7 filter |
|---|---|---|---|
| 1 | 9.00 | 25.00 | 49.00 |
| 2 | 2.019 | 2.163 | 2.205 |
| 3 | 1.471 | 1.473 | 1.471 |
| 4 | 1.314 | 1.316 | 1.316 |
| 5 | 1.238 | 1.240 | 1.240 |
| 6 | 1.192 | 1.193 | 1.194 |
| 7 | 1.161 | 1.162 | 1.162 |
| 8 | 1.139 | 1.139 | 1.140 |

Note that iterated atomic kernels $A_L^{(n)}$ are not computed in this algorithm; this definition is only used to compute the expected VRP at each iteration. Since the VRP depends on the iteration number, the filter bank needs to be pre-computed for each iteration.

---

**Algorithm 2A**: Filter bank for recursive filter.

1. For iteration $n$:
    1.1. Compute $p_{\min}^{(n)}, p_{\max}^{(n)}$ (see (22) and Table 1)
    1.2. Pre-compute lookup tables $P_n(a)$ for $0 \leq a \leq 1$ per (22).
    1.3. For discrete values of $p \in [p_{\min}^{(n)}, p_{\max}^{(n)}]$:
        1.3.1. Find $a_p = argmin(|P_n(a) - p|)$.
        1.3.2. Compute the filter bank kernel with VRP of $p$: $H_n[p] = A_L(a_p)$.
2. Output $H_n$.

---

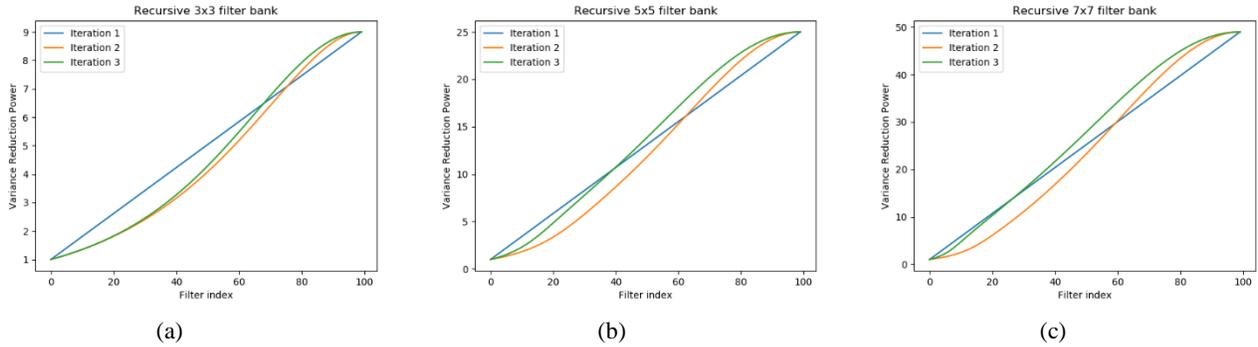

(a)     (b)     (c)

**Figure 5**. Incremental variance reduction power for 3x3, 5x5, and 7x7 recursive filter banks for the first 3 iterations. VRP for iteration 4 and higher overlap with Iteration3 plot.

Figure 5 plots VRPs (5) for 3x3, 5x5, and 7x7 recursive filter banks for the first 3 iterations. This illustration shows filter bank differences between iterations. On the other hand, after iteration 3 the filter bank does not change significantly and can be re-used for further iterations. In general, the recursive filter idea can be summarized as follows:

**Algorithm 2B**: Recursive variant filter

1. Read input arrays $f$ and $q$.
2. Pre-compute the filter bank $H_n$ (Algorithm 2A).
3. Set iteration number $n = 0$.
4. While any $q_i > q_{\min}$ repeat:
   4.1. Compute $p_{\min}^{(n)}, p_{\max}^{(n)}$ (see (22) and Table 1) and $r_{\max}^{(n)}$ for current iteration $n$.
   4.2. Loop over pixels $i \in I$ such that $q_i > q_{\min}$:
      4.2.1. Set iRVP $r = q_i$.
         If $r > r_{\max}^{(n)}$ set $r = r_{\max}^{(n)}$.
      4.2.2. Compute $p = r\, p_{\min}^{(n)}$ and select the kernel $h_p = H[p]$
      4.2.3. Convolve $\hat{f}_i^{(n)} = \hat{f}_i^{(n-1)} * h_p$.
      4.2.4. Update $q_i = q_i / r$.
   4.3. Increment the iteration number $n = n + 1$.
5. Output $\hat{f} = \hat{f}^{(n)}$.

### 3.2.1 Closed form equations for the 3x3 atomic kernel

For the simplest atomic kernel $A_1$ (11), we can derive closed-form equations for computing the 3x3 filter bank, which eliminates the need for precomputed lookup tables. We provide equations for the first three iterations; as our numerical results show, the filter bank remains almost constant after the third iteration and can be re-used. From (22), the VRP of 3x3 atomic kernel at the first 3 iterations is computed by:

$$p_a^{(1)} = \frac{(1+2a)^4}{(1+2a^2)^2} \quad (23)$$

$$p_a^{(2)} = \frac{9^2(1+2a)^4}{(3+8a+8a^2)^2} \quad (24)$$

$$p_a^{(3)} = \frac{81^2(1+2a)^4}{(19+64a+58a^2)^2} \quad (25)$$

Solving the 4$^{th}$ order polynomial equation for $a$, we can find a closed form solution for computing an atomic kernel parameter $a$ with given VRP $p$:

$$a_p^{(0)} = \begin{cases} 0.25, & p = 4 \\ \dfrac{-2+\sqrt{2t(3-t)}}{2(2-t)}, t = \sqrt{p}, & \text{otherwise} \end{cases} \quad (26)$$

for $1 \le p \le p_{\max}$.

$$a_p^{(1)} = -\frac{1}{2} + \sqrt{\frac{t}{8(9-2t)}}, \quad t = \sqrt{p} \quad (27)$$

for $p_{\min}^{(1)} < p \le p_{\max}^{(1)}$.

$$a_p^{(2)} = -\frac{32t - 162 + \sqrt{6t(81-13t)}}{324 - 58t}, t = \sqrt{p}, \quad (28)$$

for $p_{\min}^{(2)} < p \le p_{\max}^{(2)}$.

**Algorithm 2C**: Filter bank for 3x3 recursive filter

1. For iteration $n < 3$:
   1.1. Compute $p_{\min}^{(n)}, p_{\max}^{(n)}$ (see (22) and Table 1)
   1.2. For discrete values of $p \in [p_{\min}^{(n)}, p_{\max}^{(n)}]$:
      1.2.1. Compute $a_p^{(n)}$ per (26)-(28).
      1.2.2. Compute the filter bank kernel with VRP of $p$: $H_n[p] = A_L(a_p)$.
2. For $n \ge 3$ copy $H_n = H_2$
3. Output $H_n$.

### 3.2.2 Test 1 with recursive filters

To verify our implementation, we repeat the Test 1 with 3x3, 5x5, and 7x7 recursive filters. Figure 6 shows comparison of 7x7 and 11x11 fixed-size filters with the 5x5 recursive filter. As expected, recursive filter power is not limited by the kernel size, and the filtered sample variance is very close to the target value for the full range of input variance. Similar results on the Test 1 were obtained with 3x3 and 7x7 recursive filters. Figure 7 shows the number of iterations as a function of target variance reduction ratio.

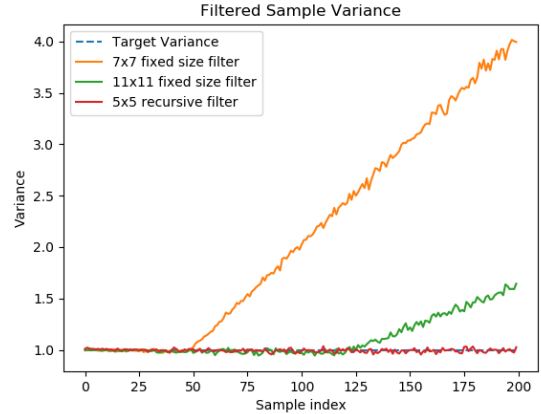

**Figure 6**. Test 1 with 5x5 recursive filter. Results show desired alignment of the filtered variance to the target values with the 5x5 recursive filter. Other tested recursive filters (3x3 and 7x7) demonstrated similar results.

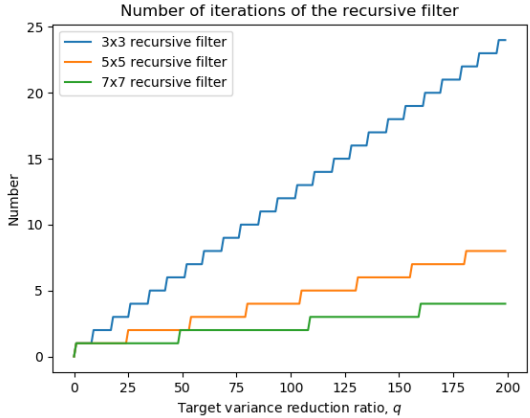

**Figure 7**. Number of iterations for 3x3, 5x5, and 7x7 recursive filters as a function of target variance reduction ratio.

#### 3.2.3  Test 2: Poisson distribution

Poisson distribution is used widely in imaging to describe data statistics [14]. In this test we simulate $N_k =100$ Poisson distributed noise samples to evaluate variance reduction performance of the proposed approach. Similar to Test 1, sample size is 128x128, and each sample $k$ is generated with linearly increasing Poisson parameter $\lambda_k \in [\lambda_{min}, \lambda_{max}]$, with $\lambda_{min} = 10$, $\lambda_{max} = 1000$. Simulated imaging task requires taking logarithm, and the target variance is given for the after-log domain. The expected mean and variance of the samples before log are given by $m_k = v_k = \lambda_k$, and the after-log variance can be estimated by the formula: $u_k = \frac{v_k}{m_k^2} = \frac{1}{m_k}$. The target variance for each sample is set to $u_T = 1/\lambda_{max}$. Log transform results in low-count samples having significantly higher after-log standard deviation than high-count samples.

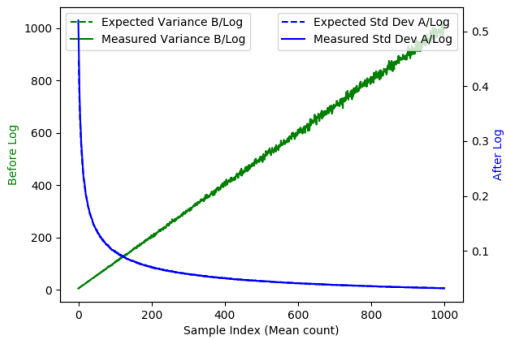

**Figure 8**. Before-log and after-log noise sample statistics with Poisson distribution. Note that before-log variance is plotted along the left vertical axis, and after-log standard deviation is plotted along the right vertical axis.

Figure 8 shows expected and measured variance of noise samples before log, as well as expected and measured standard deviation of noise samples after log. Note that filtering step is applied before log to preserve sample means.

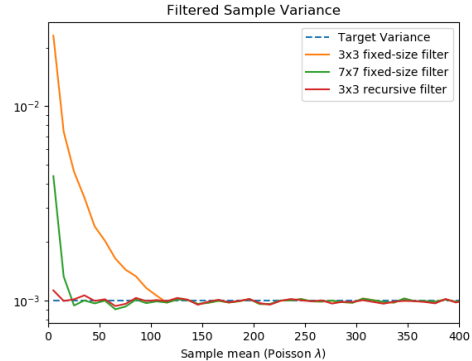

**Figure 9**. Filtered sample after-log variance for Test 2 data. Fixed-size filters deviate from target variance. Recursive filter shows good accuracy even with smallest 3x3 kernels.

Figure 9 shows after-log sample variance after applying the proposed filter. It compares 3x3 and 7x7 fixed-size filters with 3x3 recursive filter. This result shows that the proposed recursive filter is capable of strong variance reduction at low Poisson counts and provides accurate result close to the desired target value.

## 4   APPLICATIONS

### 4.1  Notes on performance optimization

The key performance characteristic of the recursive filter is the kernel size. Larger size kernels provide "higher quality" filtering; however, there are two aspects to consider when increasing the kernel size. First, the number of computations goes as a square of the kernel size, $K^2$. Second, the memory access becomes less sequential. For example, a 3x3 kernel accesses only the two adjacent rows of each pixel, while 11x11 kernel needs to access up to 5 rows in each direction. Modern parallel computing devices, such as GPUs, allow massively parallel execution of small computing tasks (kernels), and it is the memory access pattern that becomes the bottleneck of the computation speed. Therefore, objective assessment of algorithm performance is a very difficult task, it largely depends on the device architecture, framework (e.g., CUDA or OpenCL), and optimizations of the algorithm (e.g., using local vs. global memory).

Filter bank size, the number of kernels in the filter bank, is not of critical importance. Modern devices have sufficient memory to store thousands of kernels without any penalty on throughput. Kernel access time during runtime also does not depend on filter bank size.

The proposed recursive filter is implemented in Python with OpenCL API [15]. No specific optimizations were done in the code. The applications below are using 3x3 filter banks.

## 4.2 Variant Adaptive Filter for Computed Tomography

Adaptive filter is used in Computed Tomography (CT) to reduce effects of photon starvation [16] [17]. We use computer simulation to evaluate the proposed recursive filter for CT. Simulated system has 1000x64 detector with 600mm radius, 50° fan angle, and 1000 views per rotation. Let $I_i$ denote the attenuated count, where $i$ is the detector index. Similar to Test 2, the after-log variance is given by $u_i = \frac{1}{I_i}$. We apply the recursive filter before log transform to achieve uniform variance in the post-log data, i.e., variance not exceeding a certain target value $u_T$. VRR is computed by $q_i = \frac{u_i}{u_T}$ (3). In the following simulation, we used parameters $I_0 = 10^6$ (non-attenuated count) for standard dose simulation and $I_0 = 10^5$ for low dose simulation, and $u_T = 0.001$. CT data is generated for numerical shoulder phantom, with elliptical objects described in Table 3. Figure 10 shows result of variant adaptive filter application to the low dose CT data.

**TABLE 3**. NUMERICAL CHEST PHANTOM SPECIFICATION.

| Num | Center, mm | Axis, mm | Rotation, degree | HU value |
|---|---|---|---|---|
| 1 | (0, –28) | 228, 78 | 0 | 1050 |
| 2 | (0, –28) | 226, 76 | 0 | –100 |
| 3 | (0, –30) | 220, 70 | 0 | +50 |
| 4 | (160, –30) | 30, 20 | 0 | +600 |
| 5 | (–160, –30) | 30, 20 | 0 | +600 |
| 6 | (0, –60) | 20, 20 | 0 | +600 |
| 7 | (90, –50) | 30, 15 | 15 | –30 |
| 8 | (–90, –50) | 30, 15 | –15 | –30 |

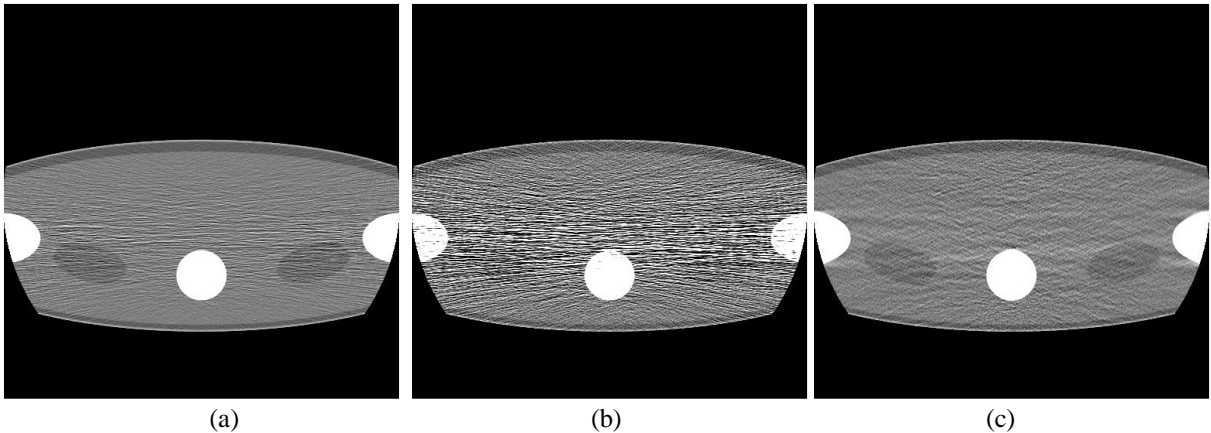

(a)     (b)     (c)

**Figure 10**. Variant adaptive filter results with CT image reconstruction. (a) Standard dose ($I_0 = 10^6$); (b) Low dose ($I_0 = 10^5$); (c) Low dose with the variant adaptive filter.

## 4.3 Edge-preserving image filter

For the edge-preserving denoising filter evaluation, we use a numerical head phantom [18] image. Edges are detected by computing the variation $\|\nabla f\|_2$, where $\nabla f$ is the image gradient. We assume a uniform input image variance, $v_0$. VRR is computed by dividing by image variation, to reduce variance reduction near the edges:

$$q = \frac{v_0}{\|\nabla f\|_2}. \quad (29)$$

An example of edge-preserving VRR is shown in Figure 11. To achieve stronger edge-preserving performance, VRR can be computed following the Perona-Malik equation [19]:

$$q = \frac{v_0}{v_0 + \|\nabla f\|_2^2} \quad (30)$$

Note that there are several approaches to estimate the input image variance, $v_0$ [20] [21] [22] [23]. The advantage of the proposed approach over the traditional edge-preserving filters (e.g., [19]), is that its variance reduction is not limited by a single iteration, and image variance does not need to be estimated at each iteration. The result of edge-preserving recursive filter with VRR given by (29) are shown in Figure 12, and results with more aggressive VRR given by (30) are shown in Figure 13.

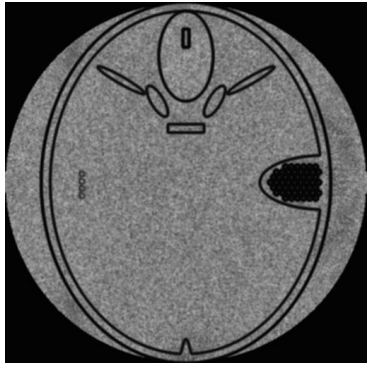

**Figure 11**. Variance reduction ratio, $q$, computed for edge-preserving denoising filter.

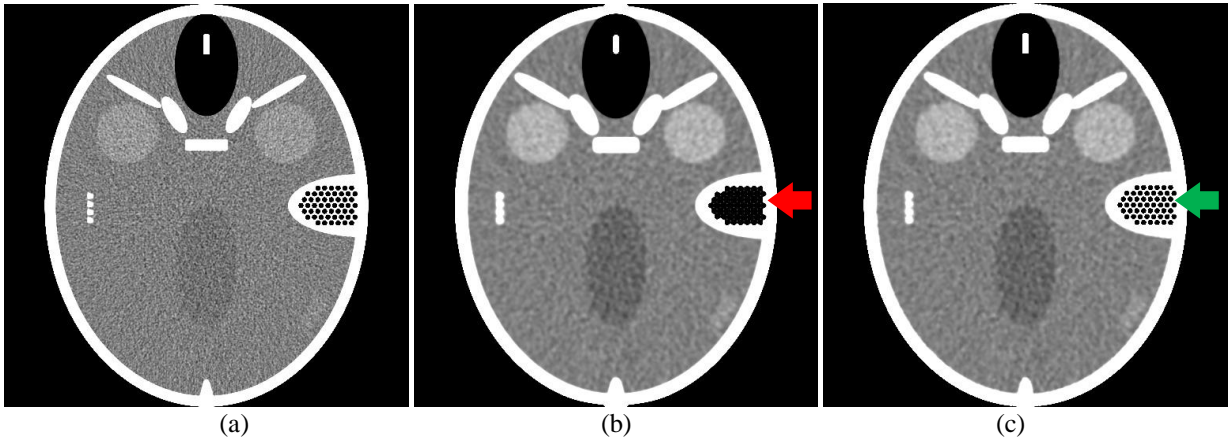

(a)            (b)            (c)

**Figure 12**. Application of recursive filter for edge-preserving noise reduction. (a) Input image (noise StDev = 8.2 HU); (b) Fixed VRR recursive filter (noise StDev = 1.8 HU); (c) Edge-preserving (29) recursive filter (noise StDev = 1.8 HU);

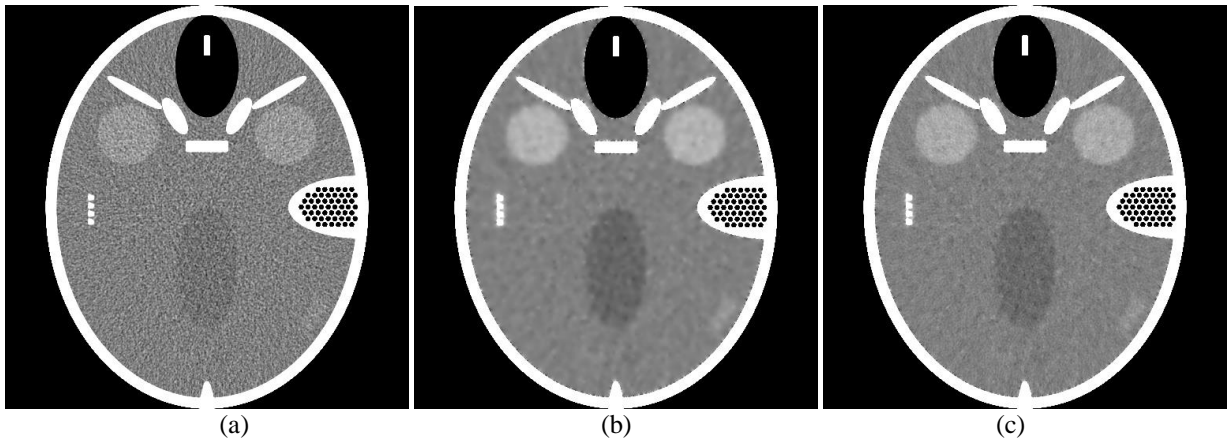

(a)            (b)            (c)

**Figure 13**. Application of recursive filter for edge-preserving noise reduction. (a) Input image (noise StDev = 8.2 HU); (b) Edge-preserving (30) recursive filter (noise StDev = 0.9 HU); (c) Combined 20% original and 80% filtered image (b) (noise StDev = 1.9 HU); Display window level and width are set to [0, 60].

## 5   CONCLUSIONS

The proposed method reduces non-uniform sample variance to the pre-determined target values, provided that the input variance can be estimated. The filter can equalize variance of the non-stationary signal, or vary filtering strength based on image features, such as edges, etc., as shown by applications in Section 4. The proposed space-variant filter approach has the following features:

- Good accuracy of variance reduction assessed at the desired target variance level.

- Strong variance reduction power, not limited by the kernel size.
- Small kernel footprint, suitable for parallel implementation.

Another novel feature of the proposed approach is the formulation of atomic kernels, that simplifies the structure of the algorithm. For the simplest form of atomic kernels, a closed-form solution was found. Atomic kernels have the following advantages:

- Continuously span the kernel space from delta kernel to the box kernel, including both.
- For any atomic kernel ($a < 1$), there is a corresponding truncated Gaussian kernel (12) with exactly same coefficients.
- Faster computation of kernel coefficients than Gaussian kernel (compare (9) and (10)).
- Kernel sum (or sum of squares) can be computed fast from generating kernel $U_L$, reducing the 2D summation ($O(K^2)$ operations) to 1D ($O(K)$ operations).

Note that the concept of atomic kernels is very general, and atomic kernels can be used anywhere the traditional Gaussian filters are used to simplify analysis of variance reduction and reduce computation time. Currently we are investigating *anisotropic* atomic kernels in the form of $U_L(a) \otimes U_L(b)$.

## REFERENCES

[1] E. Feibush, M. Levoy and R. Cook, "Synthetic Texturing Using Digital Filters," *Computer Graphics (Proc. SIGGRAPH '80),* vol. 14, no. 3, pp. 294-301, 1980.

[2] Z. Wang and A. C. Bovik, "Foveated image and video coding," in *Digital Video, Image Quality and Perceptual Coding, H. R. Wu and K. R. Rao Eds.*, Boca Raton, FL, CRC Press, 2006, p. 431–457.

[3] F. C. Crow, "Summed-area tables for texture mapping," *Computer Graphics,* vol. 18, no. 3, pp. 207-212, 1984.

[4] P. Heckbert, "Filtering by repeated integration," *ACM SIGGRAPH Comput.,* vol. 20, no. 4, p. 315–321, 1986.

[5] M. Hussein, F. Porikli and L. Davis, "Kernel integral images: A framework for fast non-uniform filtering," in *IEEE Computer Vision and Pattern Recognition*, 2008.

[6] E. Elboher and M. Werman, "Efficient and accurate Gaussian image filtering using running sums," *arXiv: Computer Vision and Pattern Recognition,* pp. 897-902, 2012.

[7] E. Elboher and M. Werman, "Cosine integral images for fast spatial and range filtering," in *18th IEEE International Conference on Image Processing*, Brussels, pp. 89-92, 2011.

[8] D. Charalampidis, "Recursive Implementation of the Gaussian Filter Using Truncated Cosine Functions," *IEEE Transactions on Signal Processing,* vol. 64, no. 14, pp. 3554-3565, 2016.

[9] R. Deriche, "Recursively implementing the gaussian and its derivatives," Tech. Rep. Res. Rep. INRIA, Sophia Antipolis, France, 1993.

[10] I. Young and L. v. Vliet, "Recursive implementation of the Gaussian filter," *Signal Processing,* vol. 44, p. 139–151, 1995.

[11] D. Hale, "Recursive Gaussian filters," CWP Report 546, Colorado School of Mines, Golden, CO, 2006.

[12] M. McDonnell, "Box-filtering techniques," *Comput. Graph. Image Process.,* vol. 17, no. 1, p. 65–70, 1981.

[13] T. Popkin, A. Cavallaro and D. Hands, "Accurate and Efficient Method for Smoothly Space-Variant Gaussian Blurring," *IEEE Transactions on Image Processing,* vol. 19, no. 5, pp. 1362-1370, 2010.

[14] J. R. Janesick, Photon Transfer, SPIE Press, 2007.

[15] A. Kloeckner, "PyOpenCL Documentation," 2009. [Online]. Available: https://documen.tician.de/pyopencl/.

[16] J. Hsieh, "Adaptive streak artifact reduction in computed tomography resulting from excessive x-ray photon noise," *Medical Physics,* vol. 25, no. 11, pp. 2139-2147, 1998.

[17] A. Zamyatin, Z. Yang, N. Akino and S. Nakanishi, "Streak artifact reduction in low dose computed tomography," in *IEEE Nuclear Science Symposium Conference Record*, Valencia, Spain, 2011.

[18] G. Lauritsch and H. Bruder, "Head Phantom," [Online]. Available: http://www.imp.uni-erlangen.de/phantoms/ head/head.html.

[19] P. Perona and J. Malik, "Scale-space and edge detection using anisotropic diffusion," *IEEE Transactions on Pattern Analysis and Machine Intelligence,* vol. 12, no. 7, p. 629–639, 1990.

[20] D. Shi, "Alternative noise map estimation methods for CT images," in *SPIE Medical Imaging: Medical Physics*, Lake Buena Vista, FL, 8668-113, 2013.

[21] K. Rank, M. Lendl and R. Unbehauen, "Estimation of image noise variance," *IEE Proceedings - Vision, Image and Signal Processing,* vol. 146, no. 2, pp. 80-84, 1999.

[22] L. Shen, X. Jin and Y. Xing, "A noise variance estimation approach for CT," in *SPIE Optical Engineering + Applications: Proceedings Volume 8506, Developments in X-Ray Tomography VIII*, San Diego, 85061M, 2012.

[23] Z. Yang, A. Natarajan and A. Zamyatin, "Practical noise assessment method," in *IEEE Nuclear Science Symposuim & Medical Imaging Conference*, Knoxville, TN, pp. 3005-3008, 2010.